\documentclass[runningheads]{llncs}
\usepackage{graphicx}
\usepackage[hyphenbreaks]{breakurl}
\usepackage[hyphens]{url}
\usepackage{booktabs}
\newcommand{\ra}[1]{\renewcommand{\arraystretch}{#1}}
\begin{document}
\title{The Prevalence of Errors in Machine Learning Experiments}
%
%
\author{Martin Shepperd\inst{1} \and
Yuchen Guo\inst{2} \and
Ning Li\inst{3} \and
Mahir Arzoky\inst{1} \and
Andrea Capiluppi\inst{1} \and
Steve Counsell\inst{1} \and
Giuseppe Destefanis\inst{1} \and
Stephen Swift\inst{1} \and
Allan Tucker\inst{1} \and
Leila Yousefi\inst{1} 
}
\authorrunning{M. Shepperd et al.}
%
\institute{Brunel University London, UK
\and
Xi'an Jiaotong University, China
\and
Northwestern Polytechnical University, China
}
\maketitle              
\begin{abstract}
 \emph{\textbf{Context}}: Conducting experiments is central to research machine learning research to benchmark, evaluate and compare learning algorithms.  Consequently it is important we conduct reliable, trustworthy experiments.
\newline
\emph{\textbf{Objective}}: We investigate the incidence of errors in a sample of machine learning experiments in the domain of software defect prediction.   Our focus is simple arithmetical and statistical errors. 
\newline
\emph{\textbf{Method}}:  We analyse 49 papers describing 2456 individual experimental results from a previously undertaken systematic review comparing supervised and unsupervised defect prediction classifiers.  We extract the confusion matrices and test for relevant constraints, e.g., the marginal probabilities must sum to one.  We also check for multiple statistical significance testing errors.
\newline
\emph{\textbf{Results}}: We find that a total of 22 out of 49 papers contain demonstrable errors. Of these 7 were statistical and 16 related to confusion matrix inconsistency (one paper contained both classes of error). 
\newline
\emph{\textbf{Conclusions}}: Whilst some errors may be of a relatively trivial nature, e.g., transcription errors their presence does not engender confidence. We strongly urge researchers to follow open science principles so errors can be more easily be detected and corrected, thus as a community reduce this worryingly high error rate with our computational experiments.
\end{abstract}

\section{Introduction}

In recent years there has been a proliferation in machine learning research and its deployment in a wide range of application domains.  The primary vehicle for evaluation has been empirical via experiments.  Typically an experiment seeks to assess the behaviour of learning algorithms over one or more data sets by varying the treatment and examining the response variables, e.g., predictive performance and execution time.  Unfortunately, a challenge has been to construct a consistent or even coherent picture from the many experimental results.  For instance, in software defect prediction, a major meta-analysis of results found that the single largest determinant of predictive performance was not the choice of algorithm but which research group undertook the work \cite{Shep14}.

As a side effect of a recent systematic review of supervised and unsupervised classifiers \cite{Li19} conducted by three of the authors (NL, MS and YG), we observed quality problems with a surprising proportion of studies when trying to reconstruct confusion matrices in order to obtain comparable classification performance statistics.  These involved violations of simple integrity constraints regarding the confusion matrix such as marginal probabilities not summing to one \cite{Bowe14}.  One likely driver for these kinds of errors is that computational experiments can be extremely complex, involving data pre-processing, feature subset selection, imbalanced learning, complex cross-validation design and tuning of hyper-parameters over many, often large, data sets.

In this study we explore the phenomenon of simple arithmetical and statistical errors in machine learning experiments and investigate one particular domain of classifying software components as defect or not defect-prone.  This is an active and economically important area.  For overviews see \cite{Cata09,Hall12}.  Note we do not consider the more complex challenges of appropriate experimental design and analysis \cite{Ioan05} nor the ongoing debate concerning the validity of null hypothesis significance testing (NHST) \cite{Colq14}.

The remainder of this paper is organised as follows.  The next section provides some background on error checking in experiments.  We then summarise how our data were extracted from a systematic review comparing unsupervised and supervised learners in Section \ref{Sec:SysRev}.  Next, in Section \ref{Sec:Anal} we explain how we checked for errors and the outcome of this analysis.  We conclude with a discussion of the significance of our findings, possible steps the research community might adopt and suggestions for further work.

\section{Background} \label{Sec:back}

For some time, researchers have expressed concerns about the reliability of individual experiments \cite{Earp15} from a range of causes including simple errors.  Brown and Heathers \cite{Brow17} analysed a series of empirical psychology studies for simple arithmetic errors e.g., if there are 10 participants, a proportion could not take on the value of 17\%.  They entitle their method granularity-related inconsistency of means (GRIM) and found that of 71 testable articles around half (36/71) appeared to contain at least one inconsistent mean.

An alternative approach is presented by Nuijten et al.~\cite{Nuij16} who provide an R package to assist in the checking of inferential statistics such as  $\chi^2$, ANOVA and t-tests.  However, their automated procedure requires the reporting of inferential statistics using the APA format which is not commonplace in computer science.  Nevertheless it is sobering to note that in their analysis of 250,000 p-values from psychology experiments, half of all published papers contained at least one p-value that was inconsistent with its test statistic and degrees of freedom. In 12\% of papers the error was sufficient to potentially impact the statistical conclusion.

More specific to experiments based on learning classifiers is the work by Bowes et al.~\cite{Bowe14} to reverse engineer the confusion matrix\footnote{A confusion matrix is a $2 \times 2$ contingency table where the cells represent true positives (TP), false negatives (FN), false positives (FP) and true negatives (TN) respectively.  Most classification performance statistics, e.g. precision, recall and the Matthews correlation coefficient (MCC), can be defined from this matrix.} from partial information, and check the satisfaction of various integrity constraints.  This has been extended and applied by Li et al.~\cite{Li19}.

This paper integrates the Li et al.~\cite{Li19} analysis with an additional category of error relating to performing multiple NHSTs without correction of the acceptance threshold, usually denoted $\alpha$ and conventionally set to 0.05.  However, NHST with multiple tests becomes problematic \cite{Bena17}.  When many tests are made, the probability of making at least one Type I error amongst the comparisons grows linearly with the number of tests.  Since some experiments make many tens or even hundreds of comparisons and rely on NHST as a means of inferencing this is a very real threat to experimental validity.  Therefore a correction should be made to the $\alpha$ acceptance threshold. The best known, though conservative, method is the Bonferroni correction which is $\alpha^\prime = \alpha / n$ where $n$ is the number of tests or comparisons.  More modern approaches include Benjamini-Hochberg which controls for the false discovery rate \cite{Benj95} and the Nemenyi post hoc procedure \cite{Dems06}.

\section{Systematic Review} \label{Sec:SysRev}

In a systematic review of studies comparing the performance of unsupervised and supervised learners, we (LN, MS and YC) identified 49 relevant studies that satisfy the inclusion criteria given in Table \ref{Tab:InCrit}.  An extended description can be found in \cite{Li19}.  The conduct of the review was guided by the method and principles set out by Kitchenham et al.~\cite{Kitc15}.  We were then able to use these 49 primary studies as a convenience sample to assess the error-proneness of computational experiments in machine learning.

\begin{table}[htp]
\ra{1.2}
\caption{Systematic Review Inclusion Criteria}
\begin{center}
\begin{tabular}{|p{2.5cm}|p{9.5cm}|}
\hline
Criterion & Description \\
\hline
Language & Written in English. \\
Topic & Applies at least one unsupervised learning method for predicting defect-prone software modules. \\
Availability & Full content must be available. \\
Date & January 2000 -- March 2018. \\
Reviewed & All papers that indicated some minimal peer review process, however, we observed some outlets appear on Geoff Beall's controversial `predatory' publisher list \cite{Perl18}. \\
Duplicates & Includes new (only the most recent version used when multiple reports of a single experiment) software defect prediction experiments.  We do not count re-analysis of previously published experiments.\\
Software system & Uses real data (not simulations).\\
Reporting & Sufficient detail to enable meta-analysis.\\
\hline
\end{tabular}
\end{center}
\label{Tab:InCrit}
\end{table}%

There is a one-to-many mapping from paper to result with the papers containing between 1 and 751 (median = 12) results apiece.  The papers cover 14 different unsupervised prediction techniques (e.g., Fuzzy CMeans and Fuzzy Self Organising Maps) coupled with 7 different cluster labelling techniques (e.g., distribution-based and majority voting).  The full list of papers and raw data may be found online\footnote{Our data may be retrieved from Figshare \burl{http://tiny.cc/vvvqbz}}.

\section{Analysis} \label{Sec:Anal}

We examine three questions.  First the prevalence of inconsistency errors relating to the confusion matrix.  Here we investigate result by result.  Second, we look at the occurrence of a particular statistical error relating to a failure to adjust the $\alpha$ threshold when using NHST form of statistical inference.  Here the unit is paper since the error relates to making \emph{multiple} tests.  Third, we examine the extent to which these problems co-occur and therefore the proportion of papers implicated.

\subsection{Inconsistency errors in the confusion matrix}

Our approach was to use the DConfusion tool \cite{Bowe14} to reconstruct the confusion matrix of classification performance for each result. This can often, but not always, be accomplished from a partial set of reported results.  For instance, if precision, recall and false positive rate are reported one can reconstruct the complete confusion matrix.  Of course, some cases (for us $\sim 33\%$ or $823/2456$) failed to report sufficient information so we cannot undertake consistency checking.  Incomplete reporting also hinders meta-analysis.
 
The next step is to test for six integrity constraints (expressed as rules that should be false).  If one or more rules are true then we know that there is some issue with the results as reported.  The cause could be as simple as a wrongly transcribed value to some deeper error. However, from the perspective of our analysis all we can say is there is a problem.  Note that the DConfusion tool also handles rounding errors which could lead to small differences in results and consequently the appearance of an inconsistency problem.  In total,   262 out of 2456 experimental results were inconsistent (see Tables \ref{Tab:ConfMatrixError} and \ref{Tab:ResErrCt}).

\begin{table}[htp]
\ra{1.2}
\caption{Distribution of confusion matrix errors}
\begin{center}
\begin{tabular}{|p{11cm}|r|}
\hline
Rule & Count \\
\hline
\textbf{1:} Performance metric out of range e.g., FMeasure $\notin [0,1] $ or  MCC $ \notin [-1,1]. $& 171 \\
\textbf{2:} Recomputed defect density $d$ is zero. & 7 \\ 
\textbf{3:} Recomputed $d$ differs from original reported defect percentage by more than 0.1 (i.e., we allow for small rounding errors).& 60 \\
\textbf{4:} Recomputed performance metrics differ from known original ones by more than 0.05 (i.e., we allow for small rounding errors).  NB, the rounding error ranges are computed by adding $\pm0.05$ to the original data unlike the more conservative range 0.01 used in \cite{Bowe14}. & 3 \\
\textbf{5:}  Internal consistency of the re-computed confusion matrix. & 19 \\
\textbf{6:} Other obvious reporting errors within paper e.g., the confusion matrix is inconsistent with their dataset or dataset summary statistics. & 2 \\
\hline
Total errors & 262 \\
Checkable and consistent & 1479 \\
\hline
\end{tabular}
\end{center}
\raggedright{{\footnotesize For a full explanation for all consistency rules refer to the figshare project \burl{http://tiny.cc/vvvqbz}.}}
\label{Tab:ConfMatrixError}
\end{table}%

\begin{table}[htb]
\caption{Proportions (with rounding) of inconsistently reported experimental results}
\begin{center}
\begin{tabular}{|l|r|r|}
\hline
Result & Count & \% of total \\
\hline
Inconsistent results & 262 & \ 10.7\% \\
Other results & 2194 & 89.3\% \\
\hline
(Other) Cannot check & 715 & 29.1\% \\
(Other) Can check - ok & 1479 & 60.2\% \\
\hline
Total & 2456 & 100\% \\
\hline
\end{tabular}
\end{center}
\label{Tab:ResErrCt}
\end{table}

\subsection{Failure to adjust acceptance threshold for NHST errors}

Irrespective of one's views regarding the validity of null hypothesis significance testing (NHST)  \cite{Colq14} it is demonstrably an error to set $\alpha$ at a particular level\footnote{Of 13 papers using NHST, 12 have $\alpha=0.05$ and, unusually, one study interprets $0.05 < p <0.1$ with $p=0.077$ as being `significant'.}   and then undertake multiple tests without making some adjustment to this threshold \cite{Bend01}.  A range of adjustment methods have been proposed subsequent to the classic Bonferroni method. In our sample, we noted researchers used either Benjamini-Hochberg \cite{Benj95} or the Nemenyi procedure \cite{Dems06} procedure.  In terms of assessing the experiment we are agnostic as to which is the `correct' adjustment.

The number of significance tests ranges from 1 to 2000 with the median=100.  Naturally the experiment that only undertakes a single NHST does not require to correct $\alpha$, however the remaining 12 experiments do.  Table \ref{Tab:NoAdjErrCt} summarises the results.  Note the experiment that makes partial corrections uses the Nemenyi post hoc test procedure for some analyses but not for the remaining 84 tests.  We thus still consider this as an error.  So more than half ($(6+1)/13$) of the experiments  that make use of NHST-based analysis are in error.  Since this is part of researchers' inferencing procedures e.g., to determine if classifier X is to be preferred to classifier Y, this is worrisome.

\begin{table}[htb]
\caption{Failure to adjust the acceptance threshold when performing \emph{multiple} NHSTs }
\begin{center}
\begin{tabular}{|l|r|}
\hline
Adjust? & Count  \\
\hline
No  & 6  \\
Partial & 1 \\
Yes & 5 \\
\hline
\textbf{Total} & 12 \\
\hline
\end{tabular}
\end{center}
\label{Tab:NoAdjErrCt}
\end{table}

\subsection{Do different types of error co-occur?}

Finally, we ask the question: is a paper that commits one class of error more likely to commit other types of error?  Table \ref{Tab:DependentErrors} gives the contingency table of the two types of error.  There does not seem to be much evidence that experimenters who make errors with confusion matrices are more likely to incorrectly deploy NHST. Consequently a highly disturbing 22 out of 35 papers contain demonstrable errors.  We cannot easily comment on the remaining 14 papers since they provide insufficient information for us to check the consistency confusion matrices (although one also contains NHST errors). 

\begin{table}[htb]
\caption{Co-occurrence of different classes of error by paper}
\begin{center}
\begin{tabular}{|p{4.5cm}|c|c|}
\hline
 & NHST Error & No NHST Error  \\
\hline
Inconsistent confusion matrix  & 1 & 15  \\
Consistent confusion matrix & 3  & 16 \\
Incomplete reporting &  3 & 11 \\
\hline
\end{tabular}
\end{center}
\label{Tab:DependentErrors}
\end{table}

\section{Discussion and Conclusions} \label{Sec:Disc}

In this study we have audited 49 papers describing experiments based on supervised learners for software defect prediction.  They were identified by a systematic review of research in this area.  We then checked for arithmetic errors and inconsistencies related to confusion matrices.  These are important since they form the basis for calculating most classification performance statistics; errors could therefore lead to wrong conclusions. In contrast, we also checked that experiments that made use of NHST type inferencing adjusted the significance threshold $\alpha$ when undertaking multiple tests to prevent inflation of the false discovery rate. This type of statistical error can also lead to wrong conclusions for researchers who wish to use p-values as a means of determining whether a result is significant or not.  

Obviously there are other classes of error one could check for in experiments.  We chose confusion matrix errors and failure to adjust significance testing for pragmatic reasons: they are objective and can be undertaken without access to the original data.  Nevertheless one must have concern that the true picture is likely to be worse than we have uncovered.  Moreover, 29\% of experimental studies fail to report sufficient information for us to be able to check for consistency.   Thus an overall (knowable) error rate of $\sim 45\%$   (22/49) of papers across all publication venues (which is likely to be an underestimate) does not inspire confidence in the quality of our machine learning experiments or at least our attention to detail.

In summary:
\begin{enumerate}
\item We have identified a number of inconsistencies or errors in a surprisingly high proportion of published machine learning experiments.  These may or may not be consequential but do raise some concerns about the reliability of analyses.
\item There are also a proportion of studies that have not published sufficient information for checking and there are of course other errors that are difficult to detect using the procedures at our disposal.
\item Our sample of experiments is a convenience sample and so is not necessarily representative of other areas of machine learning.
\item We strongly recommend that researchers adopt the principles of Open Science \cite{Muna17} so that data, experimental results and code are available for scrutiny.
\item It is our intention to communicate with the affected authors to highlight data analysis issues that \emph{seem} to require correction.  However, we recognise that mistakes can be made  by all of us, so error checking is a process that needs to be undertaken with civility and professionalism. 
\end{enumerate}		

Our analysis of errors in a sample of 49 machine learning experiments has uncovered some worrying findings.  Errors, both arithmetic and statistical are surprisingly commonplace with discoverable problems in almost 45\% of papers.  This appears broadly in line with similar analyses in experimental psychology \cite{Brow17,Nuij16}.  Nor do error rates appear to be much improved in the more obviously peer-reviewed literature.  We suggest three future lines of enquiry.  First, our sample is relatively small and non-random.  It would be interesting to see how other areas of machine learning research compare.  Second, the range of errors --- particularly statistical ones --- might usefully be explored.  Third, dialogue with authors might help us  better understand the nature of errors and their significance.  As we have stated our analysis of confusion matrices identifies inconsistencies but not the underlying causes.  Then we will be in a better position to answer the question: do the errors have material impact upon experimental conclusions?  

Finally, we strongly believe these findings should give additional impetus to the move to open science and publication of all research data, code and results.  When conducting complex computational experiments, errors may be hard to completely avoid; openness better helps us to detect and fix them.

\bibliographystyle{splncs04}
\bibliography{IDEAL2019}
\end{document}